\documentclass[12pt, a4paper]{article}

\usepackage[utf8]{inputenc}
\usepackage{amsmath}
\usepackage{graphicx}
\usepackage[margin=1in]{geometry}
\usepackage{hyperref}
\usepackage{times}
\usepackage{caption}
\usepackage{booktabs} 
\usepackage{array}   
\usepackage{float}   
\usepackage{authblk}
\usepackage{adjustbox}
\hypersetup{
    colorlinks=true,
    linkcolor=blue,
    filecolor=magenta,
    urlcolor=cyan,
}

\newcolumntype{L}[1]{>{\raggedright\let\newline\\\arraybackslash\hspace{0pt}}p{#1}}

\title{\textbf{Personal Care Utility (PCU):
Building the Health Infrastructure for Everyday Insight and Guidance
}}
\author[1]{Mahyar Abbasian}
\author[1]{Ramesh Jain}
\affil[1]{University of California, Irvine}
\date{\today}

\begin{document}

\maketitle

\begin{abstract}
Building on decades of success in digital infrastructure and biomedical innovation, we propose the Personal Care Utility (PCU) — a cybernetic system for lifelong health guidance. PCU is conceived as a global, AI-powered utility that continuously orchestrates multimodal data, knowledge, and services to assist individuals and populations alike. Drawing on multimodal agents, event-centric modeling, and contextual inference, it offers three essential capabilities: (1) trusted health information tailored to the individual, (2) proactive health navigation and behavior guidance, and (3) ongoing interpretation of recovery and treatment response after medical events. Unlike conventional episodic care, PCU functions as an ambient, adaptive companion — observing, interpreting, and guiding health in real time across daily life. By integrating personal sensing, experiential computing, and population-level analytics, PCU promises not only improved outcomes for individuals but also a new substrate for public health and scientific discovery. We describe the architecture, design principles, and implementation challenges of this emerging paradigm.
\end{abstract}

\section{Introduction: Personal Care Utility as a New Societal Infrastructure for Health}

Modern healthcare has achieved remarkable success in moments of crisis — with technology-rich environments like the Intensive Care Unit (ICU) offering extraordinary precision, real-time monitoring, and expert-led interventions. In the ICU, a team of professionals continuously tracks a wide array of biomarkers, interprets their trends, and delivers timely care with orchestration and rigor. Yet, this reactive strength has come at the expense of a deeper, more continuous engagement with health as it unfolds in everyday life. This limitation was first articulated in the early calls for precision and P4 medicine, which envisioned predictive, personalized, preventive, and participatory models of care that would complement traditional clinical practice \cite{mirnezami2012preparing,hood2011p4,flores2013p4}.

This imbalance is starkly captured in what we call the "8759 vs. 1" paradox: an individual spends 8759 hours each year outside the clinical setting, making decisions that shape their health — while barely an hour is spent in direct consultation with care providers. During those other hours, health is continuously influenced by behavior, environment, emotion, and social context. Yet, our existing computing systems remain fixated on the one hour, neglecting the remaining 8759.

We believe it’s time to reimagine health not as an occasional intervention, but as a continuous, everyday service — a utility as essential as electricity or water. Just as these public infrastructures quietly sustain modern life, health guidance should be ever-present — responsive, reliable, and deeply personalized. And just as power and water systems operate through layers of infrastructure — sensors, flow controllers, delivery pipes, and monitoring dashboards — our homes, bodies, and phones now carry the sensors and compute needed to support personalized health services at scale.  

The Personal Care Utility (PCU) reimagines healthcare by shifting the locus of health guidance from the clinic to the lived environment. It brings the orchestration and intelligence of the ICU into everyday life, not through bulky machines or human experts, but through distributed multimodal sensing, AI-guided interpretation, and empathetic interaction \cite{nag2019navigational,rahmani2022personal,jalali2014personicle,mohr2017personal,abbasian2025conversational,abbasian2024empathy}.

\noindent
Importantly, the PCU is not just a personal system. It is a public good — a dynamic, intelligent infrastructure that supports:

\begin{itemize}
    \item \textbf{Individuals} with personalized guidance for health, lifestyle, and illness management
    \item \textbf{Clinicians} with contextualized and continuous data about their patients
    \item \textbf{Researchers} with de-identified, event-based chronologies for scientific discovery
    \item \textbf{Public health systems} with real-time population-level insights for responsive planning
\end{itemize}

As millions of people interact with their PCUs — on phones, in homes, via wearables, or through community agents — the accumulated longitudinal, multimodal, and personalized data creates an unprecedented opportunity to improve population health, support equity, and accelerate biomedical research. This invisible utility becomes the connective tissue between personal aspiration and collective advancement, addressing long-recognized challenges of fragmented data, limited interoperability, and uneven access in digital health \cite{saberi2025data,smuck2021emerging,livieri2025gaps}.

\noindent
For the computing community, this poses a profound challenge. Building the PCU demands:

\begin{itemize}
    \item \textbf{Multimodal AI agents} that interpret human signals across sensors, speech, emotion, and behavior
    \item \textbf{Event-based models} that reflect the true nature of lived experience, not just statistical patterns
    \item \textbf{Personalization frameworks} that adapt to culture, preferences, and privacy expectations
    \item \textbf{Cybernetic feedback systems} where AI and human decision-making evolve together
    \item \textbf{Trust-centric design}, embedding empathy, transparency, and explainability into every interaction
\end{itemize}

The past few years have created a critical nexus — where advancements in artificial intelligence, ubiquitous sensing, human-centered computing, and health sciences have converged with urgent public health challenges. This alignment enables a transformative shift in how health can be understood, managed, and delivered, fulfilling early visions of data-driven, systems-based medicine \cite{hood2012revolutionizing}, extending big-data paradigms in behavioral health \cite{monteith2015big}, and culminating in the emergence of agentic, multimodal health infrastructures \cite{heydari2025anatomy}.

This paper presents the technical underpinnings of the Personal Care Utility (PCU), a distributed, multimodal system designed to support continuous, contextual, and personalized health engagement. Beyond a system design, it offers a framework for rethinking healthcare delivery — shifting from episodic interventions to everyday support, and from reactive models to proactive, goal-oriented, and user-centered health empowerment.

\section{Framing the Problem: What’s Missing Today}

Despite remarkable advances in digital health, our systems remain fundamentally reactive. As early proponents of personalized and systems medicine observed, the prevailing model of “sickcare” continues to channel resources toward downstream treatment rather than upstream prevention and vitality \cite{hood2011p4,flores2013p4,hood2012revolutionizing}. This orientation sustains a cycle of chronic disease management rather than enabling continuous well-being. To achieve true transformation, health must be reframed from the absence of disease to the presence of adaptive, data-driven vitality.

However, this transition is obstructed by several persistent limitations in today’s digital health landscape, repeatedly identified in the literature:

\begin{itemize}
    \item \textbf{Fragmentation:} The digital health ecosystem remains highly siloed, with data distributed across incompatible systems, devices, and platforms. Clinicians and individuals alike face a patchwork of wearables, mobile applications, and electronic records that rarely communicate effectively. This fragmentation prevents longitudinal insight and hinders data-driven coordination across contexts \cite{saberi2025data}.
    
    \item \textbf{Overload and Inequity:} Patients are often overwhelmed by fragmented metrics and notifications without meaningful synthesis or guidance. Reviews of digital health adoption show that cognitive load, usability issues, and limited health literacy reduce sustained engagement \cite{livieri2025gaps}. These gaps are further amplified by inequities in access and digital literacy, limiting the reach of current solutions.
    
    \item \textbf{Lack of Adaptive Feedback:} Most digital health applications collect and visualize data but rarely adapt based on outcomes. Few systems incorporate continuous feedback loops that personalize interventions through learning from user behavior. The absence of adaptive intelligence—despite established frameworks such as Just-in-Time Adaptive Interventions (JITAIs)—leaves users without evolving support \cite{nahum2018jitai,hornstein2023personalization}.
    
    \item \textbf{Engagement Fatigue:} Persistent notifications and gamified interfaces often lead to fatigue rather than sustainable engagement. Reviews of digital well-being apps highlight the need for context-sensitive, empathetic design that provides “just enough” guidance rather than constant demands for attention \cite{almoallim2022toward,meywirth2025personalized}.
    
    \item \textbf{No Orchestration Layer:} Despite advances in sensing, today’s systems lack an intelligent layer that connects disparate signals to coherent understanding and action. Foundational work on health navigation \cite{nag2019navigational}, multimodal mental health systems \cite{rahmani2022personal}, and recent advances in conversational and agentic health architectures \cite{abbasian2025conversational,meywirth2025personalized,heydari2025anatomy} highlight the emergence of orchestration frameworks that integrate multimodal data, reasoning engines, and empathetic interfaces—but such infrastructures remain largely experimental and inaccessible to most users.

    \item \textbf{Misinformation and Lack of Trusted Knowledge:} In the absence of curated, contextually relevant health guidance, individuals frequently rely on unverified online sources, social media, or AI-generated content of uncertain accuracy. This proliferation of misinformation erodes public trust, amplifies confusion, and can lead to harmful self-management decisions. Despite massive information availability, there remains no trusted, personalized infrastructure for verifying and contextualizing health knowledge \cite{chen2018health,swire2020public,do2022infodemics}.
 
\end{itemize}

Across these dimensions, the literature reveals a common absence: an integrated, adaptive, and empathetic infrastructure that links personal sensing to meaningful, personalized health guidance. While data abundance grows exponentially, intelligence remains fragmented—failing to translate everyday human experience into continuous, coordinated care.

\section{The Personal Care Utility: Reimagining Everyday Health Support}

A person’s health journey rarely begins with a diagnosis — it begins with a question. And more often than not, the answer comes not from a doctor, but from the swirl of information that surrounds them. Tragically, in most parts of the world, that information is unreliable at best, and dangerously misleading at worst. Commercial interests push miracle cures and supplements; religious authorities may promote rituals over science; and cultural norms often perpetuate outdated or harmful practices. In the absence of trusted, contextual, and comprehensible guidance, people fall back on hearsay — navigating one of the most complex, personal, and high-stakes domains of life with tools unfit for the task. This is not just an access problem; it is an information crisis — one that precedes, and often derails, meaningful care.

When internet became a common utility, many people’s first instinct when faced with a health concern is to consult Google—not a doctor. Searches like “What does this pain mean?” or “Why am I feeling dizzy?” often lead to overwhelming, impersonal, and context-free results. Platforms like WebMD and Mayo Clinic have provided enormous value by offering general health knowledge, but they are not designed for personal interpretation, ongoing dialogue, or contextual advice. They are static libraries—not intelligent companions.

\begin{figure}[H]
    \centering
    \includegraphics[width=\textwidth, 
                     trim={1cm 10cm 1cm 16cm}, 
                     clip]{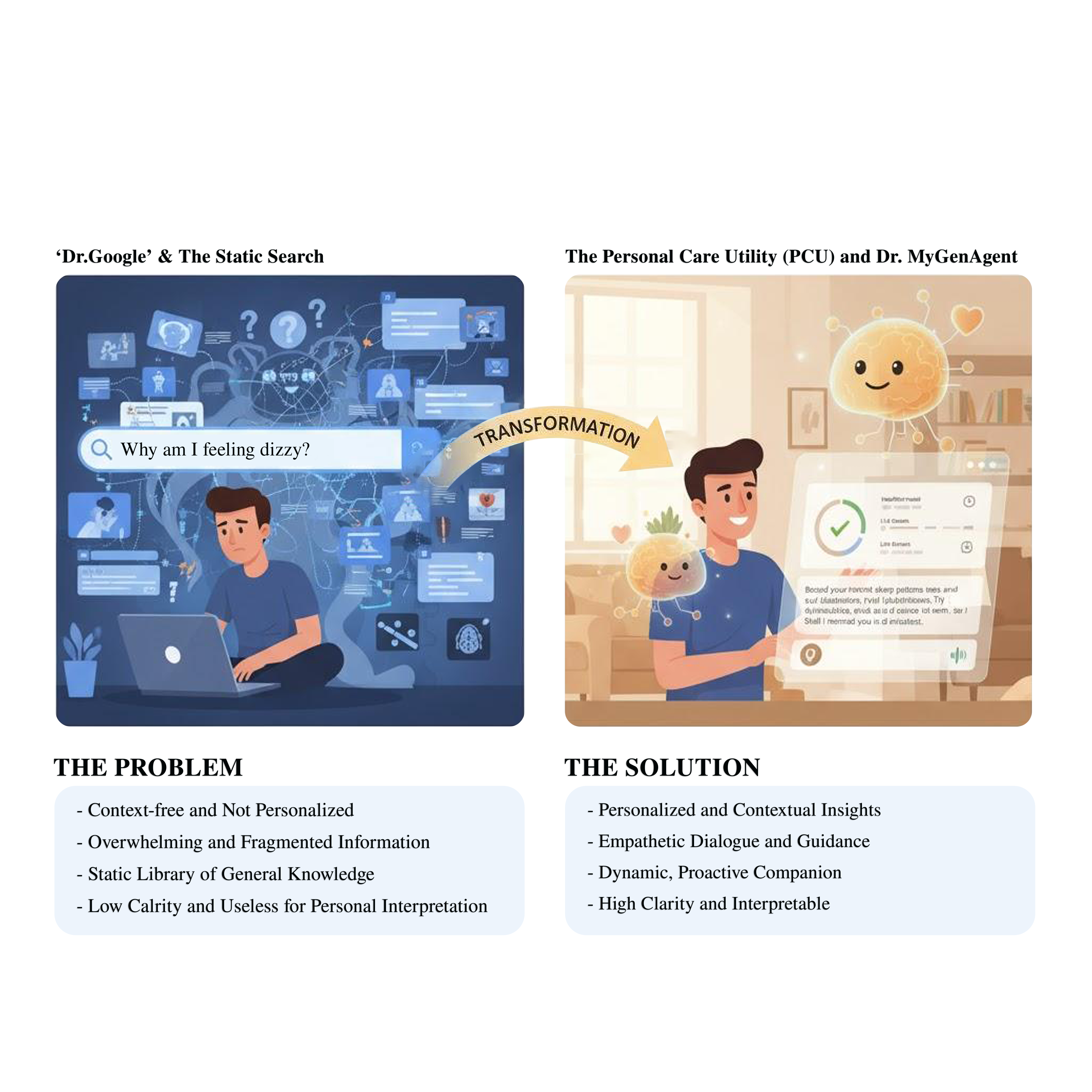}
    \caption{Transformation from static search to intelligent personal care.}
    \label{fig:drgoogle}
\end{figure}

This is where the Personal Care Utility (PCU) enters as a transformative force. At its core lies “Dr. MyGenAgent”—a personalized, always-available, multimodal health advisor built using the latest breakthroughs in AI, sensing, and contextual computing. It offers a quantum leap beyond “Dr. Google” (see Figure \ref{fig:drgoogle}).

Dr. MyGenAgent listens, observes, explains, and guides. It brings together evidence-based medical knowledge, personal context, and conversational multimodal interaction to offer clear, empathetic, and actionable advice—even before clinical care is needed.
The PCU supports three pillars of everyday health empowerment:

\begin{enumerate}
    \item \textbf{Dr. MyGenAgent: A Personalized Everyday Health Advisor}

    PCU becomes your trusted digital health companion. It answers everyday health questions, interprets bodily signals, explains diagnoses and medications, and helps assess symptoms—using voice, visuals, text, or interaction. It is powered by real-time sensing, individual history, and medical knowledge bases. It makes sense of complexity in a way no search engine can.

    \item \textbf{Personal Health Navigator for Aspirational Well-being}

    PCU helps individuals pursue long-term health and vitality—not just avoid illness. It supports daily decisions about food, sleep, activity, stress, and habits by aligning them with your goals and needs. It learns from behavior and feedback and continuously adapts to nudge healthier trajectories. It is a lifelong, personalized guide toward well-being.

    \item \textbf{Intelligent Health Information System for Personal Empowerment}

    Many people struggle to understand medical documents, prescriptions, test reports, or complex advice—especially outside the clinical context. PCU explains this information in natural, accessible terms—tailored to literacy level, culture, and current situation. It helps people navigate uncertainty, make informed choices, and engage in care with clarity and confidence—even without direct coordination with providers.
\end{enumerate}

Together, these three components offer a compelling vision of computing as a personal care utility—deeply embedded in everyday life, augmenting agency, and guiding people toward flourishing.

\section{Architecture of the Personal Care Utility (PCU)}

The Personal Care Utility (PCU) is conceived as a distributed, intelligent infrastructure that brings the proven principles of Intensive Care Units (ICUs)—continuous sensing, personalized monitoring, and timely, guided intervention—into the everyday lives of people. The PCU is not a single device, app, or service; it is an orchestrated ecosystem that transforms diverse multimodal inputs into trustworthy, understandable, and actionable health information and guidance for individuals, caregivers, and care providers.

Designed as a layered system, the PCU integrates a broad spectrum of data sources, user contexts, and knowledge systems. Each layer of the architecture contributes to a seamless transformation from raw signals to context-aware health decisions and guidance. Importantly, this architecture supports multi-user and multi-agent participation—from individuals managing their daily health to professionals analyzing population trends. The system emphasizes personalization and empathy, respects privacy and cultural diversity, and adapts dynamically as health status and context evolve.

\begin{figure}[!h] 
    \centering
    \includegraphics[width=\textwidth]{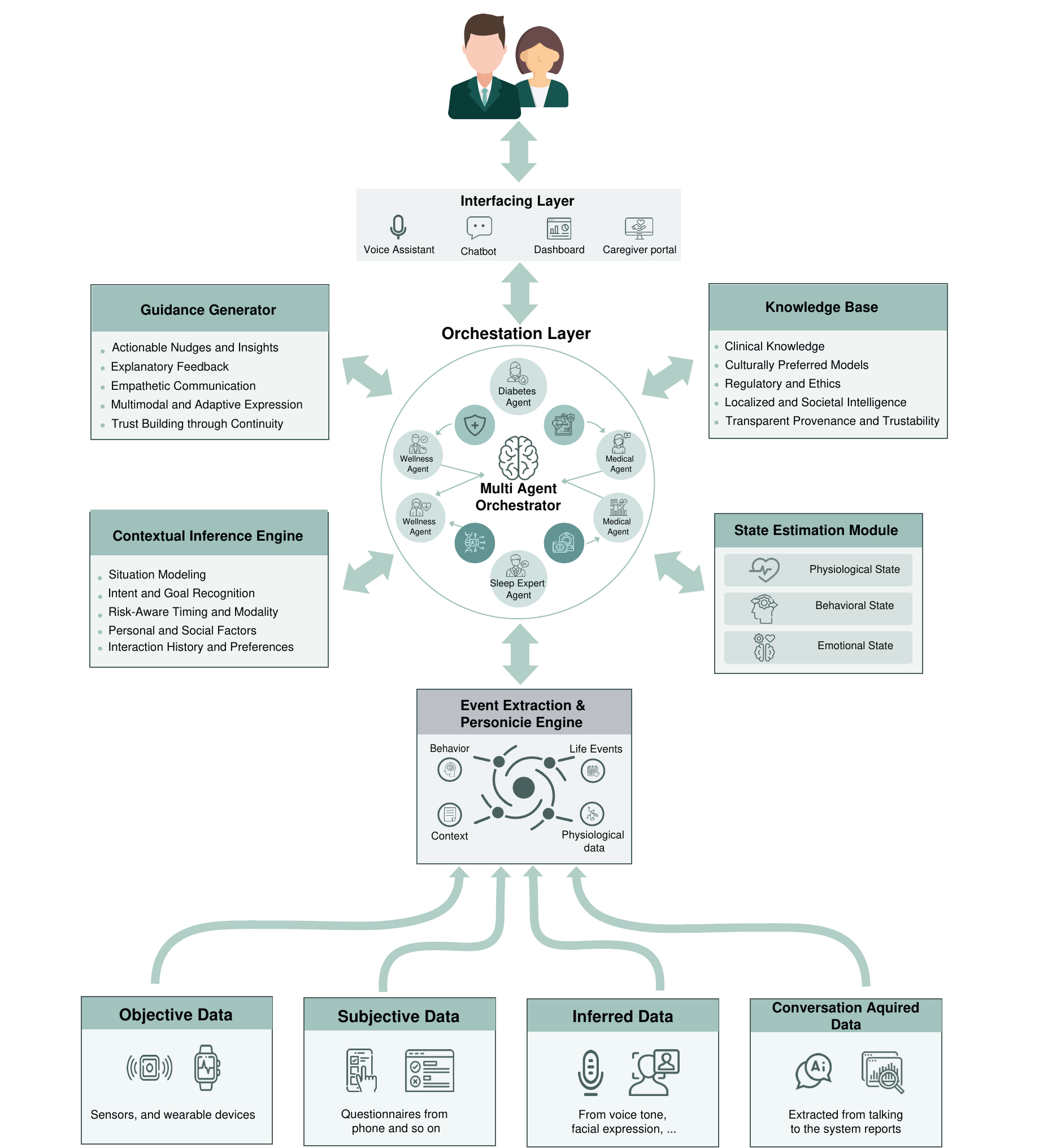} 
    \caption{Architecture of the Personal Care Utility (PCU), illustrating the system's data flow and component interactions. The system gathers Objective, Subjective, Inferred, and Conversation Acquired Data. The State Estimation Module processes these inputs, along with insights from the Event Extraction \& Personicle Engine and the Contextual Inference Engine, to continually update the user's Physiological, Behavioral, and Emotional states. At the center, the Multi Agent Orchestrator dynamically activates components—reading data from the Knowledge Base or Event engine, using the current state, and initiating actions by running specialized agents (e.g., Diabetes Agent, Wellness Agent) or the Guidance Generator. The Guidance Generator creates outputs like Actionable Nudges, delivered to users and caregivers through the Interfacing Layer. This interaction creates a feedback loop, as the system uses subsequent user inputs to determine if the guidance caused a state update or requires further attention.}
    \label{fig:pcu_architecture_final}
\end{figure}

Figure \ref{fig:pcu_architecture_final} illustrates PCU architecture, showcasing its core layers and the continuous flow of data and intelligence. This framework unifies disparate functions into a seamless service, creating a coherent data and workflow pipeline from passive sensing to personalized, proactive health interventions.

\subsection{Layer 1: Sensing Layer — Capturing the Full Spectrum of Experience}
The Sensing Layer serves as the foundation of the Personal Care Utility (PCU), gathering a diverse range of information from the physical world and user interactions. Unlike traditional health systems that rely primarily on clinical or device-generated data, PCU expands the notion of sensing into a multi-dimensional spectrum that includes:
\begin{itemize}
    \item \textbf{Objective data:} Direct measurements from wearables, mobile phones, smart home devices, and ambient sensors. This includes vital signs, activity levels, sleep patterns, and more.
    \item \textbf{Subjective data:} Self-reported observations and feelings such as pain levels, anxiety, fatigue, or mood — typically collected through simple prompts, questionnaires, or check-ins.
    \item \textbf{Inferred data:} Derived from passive observation of voice tone, facial expression, video, motion, and interaction patterns — enabling detection of emotional states or behavioral anomalies.
    \item \textbf{Conversationally acquired data:} Gathered through natural interaction with the PCU agent via text, speech, or video — allowing individuals to describe symptoms, ask questions, or reflect on their condition.
   \item \textbf{Clinical and provider-reported data:} Information originating from healthcare professionals, electronic health records, lab and imaging reports, medication lists, or structured clinical observations. These inputs provide medical validation, context for interpretation, and continuity between everyday sensing and formal care encounters.
\end{itemize}

Importantly, this architecture is inclusive by design. It supports scenarios where advanced sensors are available, but is also functional in low-resource settings where self-reporting, mobile interaction, or ambient cues may be the primary modes of sensing. This enables deployment across varied socioeconomic, cultural, and geographic contexts — from sensor-rich smart homes to rural communities.

The Sensing Layer does not assume uniformity. It recognizes variability in data quality, individual preferences, and privacy constraints — including edge-case configurations such as “ICU++” (hyper-monitoring) or “privacy-first” (minimal sensing with local-only processing). This flexibility is central to PCU's adaptability.

\subsection{Layer 2: Event Extraction and Personicle Engine — From Data Streams to Meaningful Episodes}

Once data is collected, it must be transformed into structured, meaningful representations that can support reasoning and guidance. The Event Extraction and Personicle Layer performs this transformation by organizing the sensed information into a chronological stream of life events, called a Personicle (Personal Chronicle) \cite{jalali2014personicle, jalali2021event}.

This model treats daily life not as isolated data points, but as a sequence of interpretable events \cite{oh2019detecting} such as “skipped lunch,” “slept poorly,” “took walk after meal,” “unusual heart rhythm,” or “felt anxious in meeting.” These events are semantically labeled, represented in an event model \cite{westermann2007toward}, time-stamped, and linked to context, forming the basis of personalization and explanation throughout the PCU system.

This layer differs from the common Digital Twin approach, which mirrors physiological states or system variables in real time. The Personicle instead emphasizes the narrative structure of lived experience — allowing for event-driven modeling that aligns with human reasoning, memory, and decision-making.

\noindent
Event-based models also enable:
\begin{itemize}
    \item Personal health timelines
    \item Episode-based reasoning
    \item Contextual triggering of guidance
    \item Goal-oriented planning
\end{itemize}

The Personicle is not merely a log; it is a living, evolving model of a person’s health-relevant life. It provides continuity across time and settings, and connects data to action in a way that is comprehensible to both individuals and caregivers.

\subsection{Layer 3: State Estimation Module}

This layer converts chronicle events into higher-level estimations of an individual’s current health and well-being state. Drawing from multimodal sensing, behavioral logs, and conversational inputs, it continuously infers and updates a dynamic representation of the user’s condition \cite{nag2018cross,josephine2024health}. 

\noindent The module models three broad dimensions of human state:
\begin{itemize}
    \item \textbf{Physiological State:} Heart rate variability, glucose levels, sleep cycles, fatigue, and recovery metrics.
    \item \textbf{Behavioral State:} Activity regularity, diet adherence, sleep patterns, and daily routines.
    \item \textbf{Emotional State:} Stress, anxiety, or mood inferred from voice tone, facial expressions, linguistic cues, and interaction rhythms.
\end{itemize}

\noindent It integrates multiple data sources:
\begin{itemize}
    \item \textbf{Objective:} Sensor measurements and physiological readings,
    \item \textbf{Subjective:} Self-reported pain, fatigue, or emotional states,
    \item \textbf{Inferred:} Derived from temporal and cross-modal correlations (e.g., activity–mood coupling), and
    \item \textbf{Conversational:} Information disclosed or inferred during dialogue with the user.
\end{itemize}

\noindent Each individual’s model begins with a personalized \textbf{baseline profile}—a representation of their typical physiological and behavioral ranges. This baseline evolves as more longitudinal data accumulate, allowing the system to differentiate between expected variability and true anomalies. For example, an elevated heart rate may signify stress for one person but normal exercise recovery for another.  

\noindent Importantly, this layer distinguishes between \textbf{events} and \textbf{states}. Events (as in the Personicle framework) are discrete, time-stamped observations—measurable occurrences in the user’s chronicle. In contrast, a state represents a latent, continuous construct inferred from multiple events over time. The state estimation process thus transforms fragmented event data into a coherent, contextual understanding of the person’s current condition.

\noindent This module operates continuously, balancing responsiveness with robustness by accounting for noise, missing data, and contextual ambiguity. Its output—a dynamic, multimodal state profile—serves as a crucial input to downstream reasoning and guidance layers, enabling preemptive, context-aware intervention and empathetic dialogue.

\subsection{Layer 4: Knowledge Base — Scientific, Cultural, and Regulatory Intelligence}

The Knowledge Base in the PCU serves as the evolving foundation of health guidance. Unlike static databases or narrowly scoped rule engines, it is designed as a dynamic, modular, and trustworthy repository that integrates a wide range of medical and societal knowledge — from rigorous scientific literature to regional regulations and culturally accepted practices. This integration ensures that the PCU’s reasoning and recommendations are not only evidence-based but also contextually grounded and ethically compliant \cite{abbas2025explainable,saberi2025data}.
This layer includes:

\begin{itemize}
    \item \textbf{Scientific and Clinical Knowledge:} Continuously updated repositories of clinical guidelines, biomedical ontologies, research findings, and disease models. These resources are continuously curated from trusted medical repositories and peer-reviewed literature, and may also be synthesized by AI systems to incorporate emerging evidence across diagnostics, therapeutics, and behavioral science. Together, they form the foundation for explainable and traceable reasoning within the PCU, enhancing both the interpretability and reliability of its AI-driven guidance. \cite{abbas2025explainable, gerdeskold2020use, doran2007outcomes}.
    \item \textbf{Culturally Preferred Models:} The architecture allows integration of user-selected traditional health systems such as Ayurveda, Traditional Chinese Medicine (TCM), or homeopathy through pluggable modules. When appropriate and desired by the user, these modules contribute knowledge that aligns with local beliefs and lifestyles increasing long-term adherence \cite{arshad2025personalized,kagawa2010health,guillemin1993cross,kagawa2015cultural}.
    \item \textbf{Regulatory and Ethical Constraints:} The knowledge base encodes relevant national and institutional guidelines, legal constraints (e.g., data sharing laws), and public health mandates. This ensures that PCU’s recommendations are not only personalized, but also legally and ethically compliant across different regions \cite{saberi2025data,mennella2024ethical}.
    \item \textbf{Localized and Societal Intelligence:} Region-specific practices, environmental factors, and community health insights are included to support geographically and socially contextualized care. This is essential in multilingual, multicultural, or rural environments where uniform medical models may not suffice. Such integration directly addresses barriers to accessibility and equity observed in global digital health adoption \cite{livieri2025gaps,kagawa2015cultural}.
    \item \textbf{Transparent Provenance and Trustability:} Every piece of advice or recommendation from PCU can be traced to its knowledge source — be it a WHO guideline, a peer-reviewed study, a cultural protocol, or a user preference setting. This traceability reinforces trust and accountability \cite{abbas2025explainable,guillemin1993cross,mennella2024ethical}.

\end{itemize}

Collectively, these dimensions transform the Knowledge Base into a continuously learning and ethically aware substrate for intelligent health guidance. Recent advancements in personal health agents demonstrate the feasibility of such modular knowledge integration, combining scientific, contextual, and societal intelligence into coherent reasoning pipelines \cite{heydari2025anatomy}.  In short, the Knowledge Base enables the PCU to deliver reliable, respectful, and up-to-date health guidance — grounded in evidence, aligned with user values, and adapted to real-world constraints.

\subsection{Layer 5: Contextual Inference Engine — Understanding the Situation, Not Just the Signals}

Health is not static — and neither are people’s lives. The Contextual Inference Engine serves as the PCU’s dynamic reasoning core. Its goal is to make sense of evolving situations by integrating physiological, behavioral, emotional, environmental, and social cues — then deciding when, how, and what to communicate, or whether to act at all.

This layer includes:

\begin{itemize}
    \item \textbf{Situation Modeling:} Goes beyond momentary signals to construct an evolving picture of the person’s current context — Is Raj walking to a meeting? Has Maya just finished dinner? Is this a time for focus, rest, or action? Such insights are synthesized from multimodal sources: location, calendar, personicle events, conversational tone, and physiological data \cite{oh2019detecting, jalali2014personicle,singh2012personalized,mohr2017personal}.
    \item \textbf{Intent and Goal Recognition:} Identifies the person’s likely goals or motivations in the current context — such as staying calm before a meeting, managing a sugar spike after a meal, or preparing for a social interaction. This enables goal-aligned guidance, not just raw alerts \cite{nag2019navigational,nag2018cross,nahum2018jitai}.
    \item \textbf{Risk-Aware Timing and Modality:} Determines when to intervene (or wait), and how to deliver the message (a gentle nudge via phone, a dashboard update for a caregiver, or an urgent voice alert). This is essential to avoid overloading the user or disrupting critical moments \cite{abbasian2024empathy,abbasian2025conversational,almoallim2022toward}.
    \item \textbf{Personal and Social Factors:} Recognizes social setting (alone, with family, at work), personal routines, and cultural boundaries — ensuring sensitivity in both what is said and how it is said \cite{knighton2018measuring,solomon2018health}.
    \item \textbf{Interaction History and Preferences:} Learns from past interactions to personalize further. If someone prefers visual feedback in the morning and voice reminders during walks, the system adapts accordingly \cite{oyebode2023machine, grua2022evaluation, grua2020reference}.
\end{itemize}

In essence, the Contextual Inference Engine shifts the PCU from a reactive, static, general purpose tool to a thoughtful companion — one that respects attention, timing, and personal context. It ensures that guidance is not only accurate, but also appropriate, empathetic, and well-timed.

\subsection{Layer 6: Guidance Generator — Turning Insight into Empathetic, Trustworthy Action}

The PCU is not merely a passive observer of life — it exists to help people live better. The Guidance Generator is the layer where all the upstream perception, modeling, and inference gets transformed into actionable, understandable, and personalized guidance.

\noindent
This layer delivers value by addressing three fundamental questions:
\begin{itemize}
    \item What needs to be conveyed?
    \item How should it be expressed?
    \item When is the right time to deliver it?
\end{itemize}

\noindent
Key components include:

\begin{itemize}
    \item \textbf{Actionable Nudges and Insights:} Instead of raw data or complex metrics, the system delivers simple, context-sensitive nudges, e.g., “Take a short walk in the next hour — it could help your glucose curve.” These are designed to be actionable within the person’s current setting and lifestyle \cite{orji2018persuasive, aldenaini2020trends, nahum2018jitai}.
    \item \textbf{Explanatory Feedback:} Explanations aren’t optional — they’re essential for trust. For example, “Your recent sleep pattern is affecting your energy. Here's how…” Guidance is accompanied by human-understandable reasons and optionally, supporting evidence or confidence levels \cite{oyebode2023machine,abbas2025explainable, heydari2025anatomy,rahman2025ai, saraswat2022explainable}.
    \item \textbf{Empathetic Communication:} The tone, modality, and frequency of communication are adapted to the person’s current emotional state, health condition, and social environment. When someone is anxious or frustrated, the system responds gently and supportively — not coldly or prescriptively \cite{abbasian2024empathy, abbasian2025conversational, meywirth2025personalized}.
    \item \textbf{Multimodal and Adaptive Expression:} Depending on the situation and preference, the same guidance can take different forms — a voice prompt, a visual dashboard, a text notification, or even a story-based conversational dialogue \cite{oh2019detecting, oyebode2023machine}.
    \item \textbf{Trust Building through Continuity:} Guidance is not isolated. It is designed to connect with past patterns and anticipated futures. For example: “You’ve managed your hydration well this week — nice job! Let’s keep that going today” \cite{hornstein2023personalization,grua2020reference}.
\end{itemize}

Ultimately, this layer serves as the human-facing core of the PCU. It's where users begin to feel that the system understands them, respects their individuality, and is working alongside them — not imposing rules, but guiding them with care. Empathetic, well-timed guidance isn't just about persuasion; it’s about building a sustained relationship based on trust.

\subsection{Layer 7: Orchestration Layer — AI-Powered Reasoning and Coordination}

As the Personal Care Utility (PCU) interacts with diverse individuals, caregivers, knowledge modules, and data sources, a central orchestration layer is essential. This is where reasoning, prioritization, and coordination happen — transforming raw intelligence into organized, meaningful action \cite{heydari2025anatomy,abbasian2025conversational,mesko2023imperative}.

This layer operates as a multi-agent system, where different components (agents) — such as diet advisor, sleep coach, medication tracker, emotional well-being support, or even a care team interface — function semi-independently but are centrally coordinated to ensure consistency, timeliness, and relevance \cite{milne2020effectiveness,moritz2025coordinated}.

\noindent
The core responsibilities are:

\begin{itemize}
    \item \textbf{Agent Coordination and Scheduling:} Different domains (e.g., glucose control vs. mental health) may demand attention simultaneously. The orchestration layer evaluates their urgency, context, and user state to decide what should take priority and how to communicate it.
    \item \textbf{Goal Alignment and Conflict Resolution:} Sometimes different agents might give conflicting guidance (e.g., “go for a walk” vs. “you need rest”). The orchestration layer applies contextual reasoning, goal hierarchies, and long-term personalization to resolve such conflicts intelligently.
    \item \textbf{Adaptive Reasoning and Learning:} The orchestrator doesn’t follow fixed rules. It uses AI-based reasoning and reinforcement learning to adapt based on how the user responds, evolving its coordination strategy over time.
    \item \textbf{Multi-Stakeholder Viewpoint Management:} This layer ensures different stakeholders — individuals, caregivers, professionals, even public health entities — get tailored, coordinated views of the same underlying state, without violating privacy or overwhelming users with conflicting information.
    \item \textbf{Task Decomposition and Delegation:} For more complex health journeys (e.g., post-surgical recovery), the orchestration layer breaks down overall goals into trackable, manageable subtasks assigned to appropriate modules or human actors.
    \item \textbf{Integrity, Continuity, and Trust:} Like a symphony conductor, the orchestration layer ensures that PCU behaves as one coherent system, not a chaotic set of notifications or fragmented apps. It maintains a consistent, evolving narrative that helps users trust the system and stay engaged.

\end{itemize}

This layer is where the true power of AI and contextual reasoning comes into play. It allows the PCU to behave like a lifelong companion — always learning, always adapting, and always balancing competing needs and signals to guide the user toward their long-term health and life goals.

\subsection{Layer 8: Interface Layer — Human Connection Through Empathy, Culture, and Expression}

The Interface Layer is where the Personal Care Utility becomes visible and tangible to its users. It’s not just about screens or chatbots — it’s about enabling trustworthy, empathetic, and culturally attuned interactions that feel personal, human, and helpful \cite{abbasian2025conversational, abbasian2024empathy, abbasian2024foundation}. This layer must draw from the rich traditions of experiential computing — where context, emotion, and real-world activity shape the interaction — as well as folk computing — the everyday, informal ways people use technology within their cultural and linguistic realities. By honoring how people actually live, learn, and seek help, the PCU interface can become not just a technical surface, but a companion embedded in daily life \cite{jain2003experiential, jain2003folk}.

Unlike traditional apps that present uniform outputs, the PCU interface adapts to the user’s identity, emotional state, context, and communication preferences. It must recognize that health is deeply emotional and social, not just technical. And use multimodal and cultural practices to make it very natural and friendly for a user.

\noindent
The core responsibilities are:

\begin{itemize}
    \item \textbf{Empathic Communication:} The interface doesn’t simply deliver alerts or stats — it engages in a dialogue that shows care, recognizes emotion, and responds with appropriate tone, timing, and framing. Empathy is not optional — it is core to trust and long-term engagement.
    \item \textbf{Multi-Modal Expression:} Depending on the user, context, and environment, the PCU may communicate via text, voice, video, images, stories, emojis, or visual metaphors. The goal is not just to inform but to connect, reassure, and guide.
    \item \textbf{Linguistic and Cultural Sensitivity:} Health communication must reflect local languages, idioms, metaphors, and belief systems. For example, an Ayurvedic user may respond better to expressions like “balance of doshas” than “insulin resistance.” The interface should be configurable to reflect cultural values.
    \item \textbf{Role-Based Views and Modes:} Different stakeholders — individual users, family members, doctors, community workers — may need different versions of the same reality. The interface supports role-based dashboards and privacy-aware messaging tailored to each.
    \item \textbf{Trust and Explainability:} Every suggestion or nudge should come with an optional explanation: Why now? Why this? What if I ignore it? These explanations build understanding and autonomy — crucial for lifelong engagement.
    \item \textbf{Support for Conversational Interfaces:} Conversational AI (voice or text) plays a major role. But unlike general chatbots, the PCU’s conversational agent is context-aware, emotionally tuned, and personalized — more like a health companion than a generic assistant.
    \item \textbf{Support for Accessibility and Dignity:} Interfaces must be designed with aging users, people with disabilities, and low-literacy populations in mind — emphasizing clarity, minimalism, and dignity. A good interface is not flashy — it is unobtrusive, empowering, and respectful.

\end{itemize}

In short, the Interface Layer is not just how the PCU speaks — it’s who the user believes they are interacting with. This layer is key to shaping a user’s perception of the PCU as a trusted, caring presence — not just a system, but a companion.

Together, these eight layers form an end-to-end infrastructure for real-time, personalized, and context-aware health guidance that operates across the 8,759 hours of everyday life — not just the 1 hour in a clinical setting.

\section{Multimodal AI as the Enabler}

The Personal Care Utility (PCU) is not merely a data processing system — it is an intelligent agent deeply embedded in the lived experiences of individuals. To achieve its mission of personalized, contextual, and trustworthy health support, the PCU must understand and interpret the full range of signals that characterize human life. This calls for a robust, integrated approach to Multimodal AI — one that brings together diverse data types, sensory modalities, interaction styles, and communication formats \cite{stahlschmidt2022multimodal,mohr2017personal,zhang2024passively,smuck2021emerging,aczon2024automated}
.

Multimodality in the PCU context refers to the integration of sensor data (e.g., physiological, behavioral, environmental), digital interactions (e.g., smartphone usage, app patterns), and human communication (e.g., speech, facial expressions, gestures, text conversations). Each of these modalities captures different facets of the person’s state and environment, and together they form a rich, layered tapestry from which meaning can be inferred \cite{jalali2014personicle,oh2019detecting,jalali2021event,nag2018cross,josephine2024health}.

However, integrating these modalities is non-trivial. Each signal comes with its own semantics, reliability, noise characteristics, and sampling dynamics. More importantly, their meaning often emerges only when considered in context — a smile may indicate joy or may be masking discomfort; elevated heart rate may signal exertion or stress depending on the situation. PCU’s architecture addresses this by transforming raw data into chronicles of life events (via the Personicle layer), estimating current states (physiological, emotional, behavioral), and drawing from knowledge bases and contextual inference engines to interpret and guide.

The real power of Multimodal AI lies not in individual modalities but in their coordinated interpretation — and this is where multimodal agents become essential \cite{jain2024multimodal}. These agents don’t just process signals; they orchestrate them. They act as intelligent intermediaries that can fuse diverse streams of data, adapt to user context, maintain memory across time, and reason about what action or feedback is most appropriate. As such, they become the dynamic, always-on interpreters of lived experience — allowing the PCU to act more like a sensitive, evolving companion than a collection of static tools.

\noindent
The real power of Multimodal AI lies not in individual modalities but in their coordinated interpretation. This requires:
\begin{enumerate}
    \item \textbf{Representation and Fusion:} Aligning and encoding different types of inputs in ways that preserve their unique meaning but also allow for joint reasoning.
    \item \textbf{Temporal Modeling:} Understanding sequences and co-occurrences across time (e.g., how behavior, context, and biological signals evolve over hours, days, or weeks).
    \item \textbf{Personalization:} Learning individual-specific patterns — what’s “normal” for one person may be a sign of distress for another.
\end{enumerate}

The PCU must also be a conversational agent, capable of interacting in human-centric ways \cite{abbasian2025conversational}. Whether through chat, speech, or visual dashboards, the PCU must understand multimodal input from the user (language, tone, gestures) and respond appropriately, respecting user preference, emotion, and cognitive load. While we have not yet formalized our research on “adaptive expression,” the PCU does emphasize interaction that adjusts not just to the user’s state, but also to their context, goals, and mode of interaction — a key direction for future exploration.

Finally, the PCU, as a multimodal utility, operates across a spectrum of interfaces and agents — from wearable feedback to caregiver summaries — necessitating multi-agent coordination and perspective-taking. What the user sees, what the doctor sees, and what the system internally tracks must all stay coherent, consistent, and interpretable.

Thus, the multimodal AI foundation of the PCU is central not only to its technical capabilities but also to its human relevance. It is how the PCU becomes more than a tool — it becomes a trusted companion in the health journey of individuals and communities.

\section{Personalization and Human Sensitivity}

Personalization is not an optional feature in healthcare—it is foundational for trust, empathy, and sustained engagement. For the Personal Care Utility (PCU), personalization goes beyond surface-level recommendations; it refers to the ability to tune data collection, interpretation, communication, and guidance to the individual’s preferences, conditions, abilities, and cultural norms.

To make this tractable, we frame personalization as operating at multiple levels, from minimal to intensive. These levels are not rigid stages but rather adaptive zones that respond dynamically to a person’s current health situation, preferences, and context.

To illustrate, consider two individuals: Raj and Maya.
\begin{itemize}

\item Raj is a cautious, private man in his 50s. He wears a smartwatch for tracking steps but turns off GPS and avoids sharing personal data unless absolutely necessary. He is skeptical of AI guidance and prefers consulting his longtime physician.
\item Maya, in contrast, is a vibrant woman in her early 50s who embraces technology. She shares health updates with her family via apps, uses AI-generated summaries after doctor visits, and is eager to try personalized food or exercise recommendations based on her menstrual cycle, mood, and sleep.

\end{itemize} 

For both of them to benefit from the PCU, the system must adapt — not only to their physiological differences but to their attitudes toward data, trust, culture, and control. It must know when to speak up, when to stay silent, when to ask, and when to wait.

\subsection{Five Levels of Personalization: A Developmental Framework}

To operationalize personalization, we propose a five-level model — inspired by the levels of autonomous driving\cite{on2021taxonomy} — that delineates increasing system capability and corresponding user involvement.
To help system designers conceptualize the range of interaction strategies, we present a simplified personalization framework. Table \ref{tab:personalization} is not prescriptive but illustrative—showing the importance of tuning not just the content, but also the tone, timing, and modality of interaction.

\begin{table}[h]
    \centering
\caption{Levels of personalization}
\label{tab:personalization}
\begin{adjustbox}{width=\textwidth}
    \begin{tabular}{|c|c|c|c|}\hline
         \textbf{Level}&  \textbf{Interaction Style}&  \textbf{User Engagement}& \textbf{Example Use Case}\\\hline
         0&  Generic broadcast&  No personalization& General health news app
\\\hline
         1&  Demographic personalization&  Age/gender-based messaging& Exercise tips for seniors
\\\hline
         2&  Behavioral tailoring&  Routine-aware nudges& Reminders based on sleep/activity patterns
\\\hline
         3&  Situational adaptation&  Context-aware interaction& Stress-reduction suggestions during travel
\\\hline
         4&  State-sensitive engagement&  Emotion- or crisis-sensitive& Intervention during depressive episodes
\\\hline
         5&  Co-regulated companionship&  Deep personalization + empathy& Real-time check-ins and explanations during illness flare-ups
\\ \hline
    \end{tabular}
    
\end{adjustbox}
\end{table}

These levels serve as a scaffold for:
\begin{itemize} 
\item Developers: to understand data and capability requirements.
\item Users and Clinicians: to build transparency, set expectations, and control engagement.
\item Policy Makers: to craft appropriate safeguards and user consent protocols.

\end{itemize} 

This framework does not insist that every PCU must operate at Level 5. In fact, individuals like Raj may prefer to cap the system at Level 3, while others like Maya may embrace Level 5 across most domains. Just as autonomous driving allows different levels of control, so too must health systems support graduated autonomy and personalization.

\subsection{Design Implications}
A sensitive personalization strategy requires:
\begin{itemize}
\item Progressive trust-building rather than all-or-nothing consent.
\item Explicit user control over how much the system knows and does.
\item Fallback to human support when needed.
\item Cultural sensitivity that respects beliefs, taboos, and social norms.

\end{itemize}

Above all, personalization must feel empowering, not intrusive. It should act as a mirror, not a magnifying glass; a companion, not a commander. And it must constantly learn — not only who we are, but how we wish to be known.

The PCU architecture, with its layered design, supports these personalization levels by enabling flexible orchestration of data sensing, state estimation, contextual reasoning, and guidance generation. Empathetic personalization is not just about smarter technology—it is about honoring the person at the center of care.

\section{Discussion}

The promise of the Personal Care Utility (PCU) will be judged by the reliability, usefulness, and safety of the \emph{final answers} it provides in the flow of everyday life. Because PCU responses are generated by adaptive, data-driven agents, evaluation cannot stop at offline model metrics. It must directly assess the quality of user-facing outputs, their evidentiary grounding, and their downstream impact on decisions and well-being.

\subsection{Evaluating Final Answers from PCU}

We frame answer evaluation as a layered pipeline combining automated checks with human review:

\begin{itemize}
  \item \textbf{Groundedness \& Attribution:} Each recommendation should be explicitly tied to vetted sources in the Knowledge Base, with traceable provenance and optional inline explanations. Automated scoring of grounding/attribution and explanation quality provides continuous QA, complemented by periodic expert audit \cite{abbasian2024foundation,abbas2025explainable,rahman2025ai,saraswat2022explainable}.
  \item \textbf{Clinical Correctness \& Safety:} Responses are validated against current guidelines and domain rules; high-risk scenarios trigger guardrails and graceful refusal or escalation. Clinician-in-the-loop review of representative vignettes and prospective spot checks reinforce safety beyond automated tests \cite{abbas2025explainable,mesko2023imperative,gerdeskold2020use}.
  \item \textbf{Faithfulness \& Consistency Over Time:} The system monitors contradiction rates across sessions, verifies that advice stays consistent with user state and medications, and calibrates confidence displays. Longitudinal “drift” dashboards help detect degrading answer quality.
  \item \textbf{Human Factors: Empathy \& Comprehensibility:} Beyond being correct, answers must be understandable and supportive. PCU tracks clarity, tone, and perceived empathy via lightweight user ratings and rubric-based reviews for conversational quality \cite{abbasian2024empathy,abbasian2025conversational, milne2020effectiveness}.
  \item \textbf{Outcome-Linked Learning:} Where appropriate, anonymized, consented feedback loops tie answer patterns to proximal behaviors (e.g., adherence to nudges) and longer-term trajectories. This complements just-in-time adaptation with evidence of real-world utility \cite{nahum2018jitai,grua2022evaluation,oyebode2023machine}.
\end{itemize}

Together, these elements move evaluation beyond static benchmarks to \emph{living assurance}: continuously measured groundedness, safety, empathy, and usefulness in place \cite{abbasian2024foundation,abbas2025explainable}.

\subsection{Privacy, Autonomy, and Governance}

PCU treats privacy as a design primitive. Data collection is \emph{purpose-bound and minimal}, with options for on-device processing and selective sharing. Federated and privacy-preserving learning strategies can reduce central aggregation and exposure, while transparent consent and role-based views respect personal, familial, and clinical boundaries. Interoperability is implemented with patient-centric controls, not just technical bridges \cite{saberi2025data}. Ethical and regulatory expectations for AI in health—transparency, accountability, bias mitigation, and auditability—are operationalized as runtime policies inside the orchestration and knowledge layers \cite{mennella2024ethical,mesko2023imperative}. Culturally aware defaults and communication further reinforce dignity and trust across communities \cite{kagawa2015cultural}.

\subsection{A Positive, Data-Rich Future}

The near future is unmistakably multimodal: wearables, ambient sensors, and passive digital traces will make high-frequency health signals commonplace \cite{smuck2021emerging,mohr2017personal,zhang2024passively}. With responsible governance, this abundance becomes an asset—fueling better state estimation, earlier detection, and more personalized coaching—while multimodal fusion in clinical data streams improves prediction and coordination of care \cite{aczon2024automated}. As coordinated agentic systems mature, the PCU can keep user, caregiver, and clinician views coherent and comprehensible, turning raw data into compassionate, timely guidance at scale \cite{heydari2025anatomy,moritz2025coordinated}.

\section{Conclusion}

This paper introduced the Personal Care Utility (PCU): a distributed, multimodal infrastructure for continuous health guidance across everyday life. We articulated an end-to-end architecture—from sensing and event extraction (Personicle) to state estimation, knowledge integration, contextual inference, guidance generation, orchestration, and interface design—that operationalizes a shift from episodic interventions to sustained, context-aware support. By aligning multimodal agents with event-centric modeling and empathy-aware interaction, PCU provides a technical substrate for delivering individualized information, proactive navigation, and post-event recovery support at scale.

A central contribution is the integration of evaluation and governance into the system design. We outlined a “final-answer” assessment pipeline that couples automated groundedness, safety checks, and longitudinal consistency monitoring with human review, and connects evaluation to outcomes where appropriate. Privacy, autonomy, and cultural sensitivity are treated as design primitives through purpose-bounded data collection, selective sharing, and role-based views, complemented by provenance-aware knowledge integration.

PCU’s potential rests on near-term trends in multimodal sensing, on-device learning, and agentic coordination. Realizing this potential requires addressing open challenges: (i) robust fusion under heterogeneous data quality and missingness; (ii) scalable personalization with explicit user control; (iii) prospective clinical and population-level studies that quantify benefit, risk, and equity impacts; and (iv) operational frameworks for auditing bias, drift, and safety in the wild. These constitute a concrete agenda for the computing and health communities.

In sum, PCU reframes everyday environments as computable contexts for trustworthy health guidance. By embedding evaluation, privacy, and cultural alignment into its core, the framework provides a path toward reliable, equitable, and empathetic health support across the 8,759 hours outside the clinic. The next step is rigorous, real-world validation and iterative refinement in partnership with clinicians, communities, and policymakers.

\bibliographystyle{ieeetr}
\bibliography{references}

\end{document}